\documentclass[11pt,a4paper]{article}

\usepackage[utf8]{inputenc}
\usepackage[T1]{fontenc}
\usepackage{lmodern}
\usepackage[a4paper,margin=2.5cm]{geometry}
\usepackage{amsmath,amssymb}
\usepackage{graphicx}
\usepackage{booktabs}
\usepackage{caption}
\usepackage{xcolor}
\usepackage{enumitem}
\usepackage[hidelinks]{hyperref}
\usepackage{microtype}
\usepackage{placeins}

\title{\textbf{Where Does the Semantic Gain Come From?}\\[4pt]
\large A Reproduction and Extension of Semantic Knowledge-driven
Contrastive Learning for Long-Tailed Recognition}

\author{Sushrut Ghimire\\[2pt]
\small Friedrich-Alexander-Universit\"at Erlangen-N\"urnberg (FAU), Germany\\
\small \texttt{sushrut.ghimire@fau.de}}

\date{}

\begin{document}
\maketitle

\begin{abstract}
Semantic Knowledge-driven Contrastive Learning (SKCL) uses a language model to
decide which classes are related, and pulls each image towards the prototypes
of its semantic neighbours. On CIFAR-100-LT ($\beta = 100$) it reports $54.02\%$
top-1 accuracy, $2.01$ points above Balanced Contrastive Learning (BCL), the
method it builds on. The code and the class descriptions are not public. I
reimplement SKCL, BCL and ConCutMix in one framework, check it against the
public baseline code, and run every configuration with three seeds. The two
baselines reproduce within $1.5$ points, but SKCL built on BCL, as the paper
describes it, ends up $0.56$ points \emph{below} BCL. To find out why, I add
SKCL to the authors' own ConCutMix code. Trained for the paper's $300$ epochs,
it reaches $53.69$, only $0.33$ below the published number. At the same budget,
however, the semantic graph adds just $0.23$ points over ConCutMix, while
training ConCutMix for $100$ more epochs adds $1.04$. Together with ConCutMix's
published lead over BCL ($1.15$), this explains the claimed gain. A BCL model
that never sees the graph already shares $41.2\%$ of the graph's top-2
neighbours with its own most-confused classes ($2.0\%$ by chance), which shows
why the graph adds so little on these benchmarks. I also test several changes
to SKCL. Combining it with the CutMix branch improves it by $1.06$ points, and
an adaptive version of the graph improves it slightly ($+0.34$ and $+0.28$ in
two codebases), although these gains are within seed noise.
\end{abstract}

\section{Introduction}

In long-tailed recognition, a few classes have many training images and most
classes have very few. A tail class with five images does not give enough
information to learn where it should lie in feature space. One natural idea is
to let rare classes borrow from related classes that have more data. The
difficulty is to know which classes are related.

SKCL~\cite{liu2025skcl} answers this with a language model. It writes a short
description of every class, embeds the descriptions with a sentence encoder,
and turns their similarities into a graph over fine classes and superclasses.
The contrastive loss then pulls each image towards its own class prototype and
also towards the prototypes of its nearest classes in this graph. The idea is
cheap, since it needs only one sentence per class, and it brings in knowledge
from outside the imbalanced image data. On CIFAR-100-LT the paper reports
$54.02\%$, which is $2.01$ points above BCL~\cite{zhu2022bcl} and $0.86$ points
above ConCutMix~\cite{pan2024concutmix}.

The method rests on an assumption that the paper does not test: that the
relations language knows are relations the vision model would not find by
itself. This matters, because if the network already confuses otters with seals,
being told that they are similar does not help. There is also a practical
problem. No code and no class descriptions are released, so the input of the
method, which is the language-model output, cannot be checked.

In this paper I reproduce SKCL and then try to understand where its gain comes
from. My main findings are:

\begin{itemize}[leftmargin=1.4em,itemsep=2pt]
  \item Built on BCL, as described in the paper, SKCL does not improve over BCL
        ($-0.56$ points, three seeds), although BCL and ConCutMix themselves
        reproduce within $1.5$ points of their published values.
  \item Built on the authors' own ConCutMix code and trained for $300$ epochs,
        SKCL reaches $53.69$, within $0.33$ points of the paper. So the result
        can be reproduced, but in a different setting than the paper describes.
  \item The claimed $+2.01$ over BCL can be explained by the ConCutMix branch
        ($+1.15$) and a longer schedule ($+1.04$), while the graph adds about
        $0.2$ points at equal budget.
  \item The graph is mostly redundant on CIFAR. A model that never sees it
        already finds $41.2\%$ of its top-2 edges ($2.0\%$ by chance), and
        changing which edges the graph contains does not change accuracy
        beyond noise.
  \item Some changes do improve SKCL, mainly combining it with the CutMix
        branch ($+1.06$). An adaptive graph that mixes the language graph with
        the classes the network has learned gives small gains ($+0.34$,
        $+0.28$) that are within noise.
\end{itemize}

Along the way I found eight differences between my first implementation and
the official baseline code, two reproducibility bugs in my own code, and three
problems that stop the official ConCutMix script from running. I describe these
because they are the kind of details that decide whether a reproduction
succeeds.

\section{Background}

\textbf{Long-tailed recognition.} Methods can be grouped into data-level,
loss-level and representation-level methods. Logit adjustment~\cite{menon2021logit}
adds $\tau\log\pi_c$ to the logit of class $c$ during training and is used by
all methods in this paper. CutMix~\cite{yun2019cutmix} pastes a patch of one
image into another. ConCutMix~\cite{pan2024concutmix} adapts this to long-tailed
data and corrects the mixed label using similarities in a learned feature space.

\textbf{Balanced contrastive learning.} Supervised contrastive
learning~\cite{khosla2020supcon} is biased under imbalance because head classes
dominate both anchors and negatives. BCL~\cite{zhu2022bcl} fixes this with
\emph{class-averaging}, which averages negatives within each class, and
\emph{class-complement}, which adds one learnable prototype per class to every
batch so that every class is present.

\textbf{SKCL.} For each fine class $i$, SKCL defines a member set $M_i$, which
is the class itself plus its top-$k$ most similar classes in the language graph
$S$. With unit-norm embeddings $z_i$ and prototypes $c_q$, its first loss term is
\begin{equation}
\mathcal{L}^{(1)}_i = -\frac{1}{|M_i|}\sum_{c_q \in M_i}
\log \frac{\exp(z_i^\top c_q / \tau')}{\sum_{n} \exp(z_i^\top c_n / \tau')} ,
\label{eq:skcl1}
\end{equation}
where the sum in the denominator runs over all fine and coarse prototypes. The
second term is BCL's class-averaged contrastive loss. If $M_i$ contains only the
own class and the coarse head is removed, SKCL becomes BCL. This makes the
comparison between SKCL and BCL a clean test of the graph.

\section{Reproduction Setup}

\textbf{Data.} I use CIFAR-10-LT and CIFAR-100-LT with imbalance factor
$\beta = 100$, built in the standard way~\cite{cao2019ldam}. CIFAR-100-LT goes
from $500$ to $5$ images per class ($10{,}847$ images). The test sets stay
balanced. Classes are grouped into Many ($>100$ images, $35$ classes), Medium
($20$--$100$, $35$) and Few ($<20$, $30$).

\textbf{Graph.} I generated one description per class with an LLM, using the
template given in the paper, for the $100$ fine and $20$ coarse classes. The
descriptions are embedded with \texttt{all-MiniLM-L6-v2}~\cite{reimers2019sbert}
and the graph is their cosine similarity matrix. I use $k = 2$. The strongest
links make sense, for example \textsf{chair}--\textsf{table} ($0.73$) or
\textsf{otter}--\textsf{seal} ($0.57$).

\textbf{Model and training.} All three methods share a ResNet-32 with cosine
classifiers, a projection head ($64 \to 512 \to 128$) and a prototype head that
maps classifier weights to prototypes. Each image gives one view for the
classifier (crop, flip, AutoAugment~\cite{cubuk2019autoaugment},
Cutout~\cite{devries2017cutout}) and two contrastive views. Training uses SGD
with learning rate $0.1$, cosine decay, batch size $256$ and $300$ epochs for
every method. The temperatures are $\tau = 0.1$ and $\tau' = 0.2$, and the
coarse classification loss has weight $0.5$.

\textbf{Calibration against the official code.} My first version was $4$--$5$
points below the published values for \emph{all} methods. Because the gap was the
same everywhere, the problem had to be in the shared code. Comparing line by line
with the public BCL and ConCutMix repositories, I found eight differences. The
biggest were missing Cutout ($+2.10$ points when fixed) and applying CutMix to
every batch instead of half of them ($+2.89$).

\textbf{Two reproducibility bugs.} Two runs with the same seed first differed by
$0.61$ points. One reason was that Cutout drew its random position from NumPy's
global state inside forked data-loader workers, which PyTorch does not reseed,
so the seed did not control Cutout. The other was that the prototype head used
batch normalisation over the \emph{classes}, which is only two rows for the
CIFAR-10 coarse level. After fixing both (per-worker seeding and layer
normalisation) the same seed gives bit-identical runs. All results below come
from the corrected pipeline.

\textbf{Metric.} There is no validation set, so reporting the best epoch means
selecting on the test set. In my framework I therefore report the mean top-1
accuracy over the last ten epochs, averaged over three seeds. On the official
ConCutMix code I also report the best epoch, because this is how the original
numbers are reported.

\section{Results}

\subsection{SKCL built on BCL does not reproduce}

\begin{figure}[t]
\centering
\includegraphics[width=0.92\textwidth]{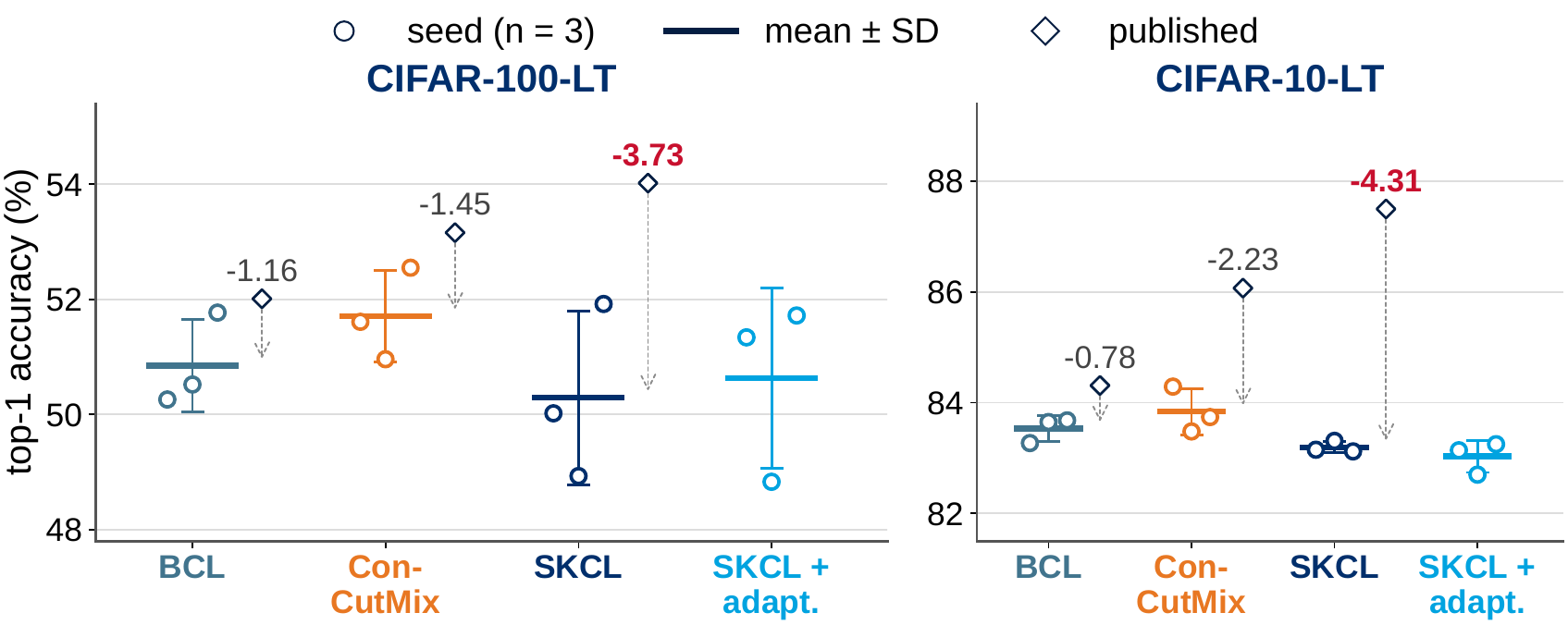}
\caption{Per-seed results (open circles), mean $\pm$ SD (bars) and published
values (diamonds). Numbers give ours minus published. SKCL + adapt.\ is the
adaptive graph of Section~\ref{sec:improve}.}
\label{fig:main}
\end{figure}

Table~\ref{tab:main} and Figure~\ref{fig:main} show the main result. BCL and
ConCutMix are $1.16$ and $1.45$ points below their published values on
CIFAR-100-LT, which is within the range one expects from a reimplementation.
SKCL is $3.73$ points below, and it is $0.56$ points below BCL, with per-seed
differences of $-0.24$, $-1.59$ and $+0.15$. The paper claims $+2.01$. On
CIFAR-10-LT the picture is the same: SKCL is $0.34$ below BCL, where the paper
claims $+3.19$.

\begin{table}[t]
\centering
\small
\caption{Top-1 accuracy (\%), $\beta = 100$, ResNet-32, my framework; mean
$\pm$ SD over three seeds of the last-10-epoch mean. Published values
from~\cite{liu2025skcl}.}
\label{tab:main}
\begin{tabular}{@{}lcccc@{}}
\toprule
& \multicolumn{2}{c}{CIFAR-100-LT} & \multicolumn{2}{c}{CIFAR-10-LT} \\
\cmidrule(lr){2-3}\cmidrule(l){4-5}
Method & Ours & Published & Ours & Published \\
\midrule
BCL       & $50.85 \pm 0.81$ & 52.01 & $83.53 \pm 0.23$ & 84.31 \\
ConCutMix & $51.71 \pm 0.80$ & 53.16 & $83.84 \pm 0.41$ & 86.07 \\
SKCL      & $50.29 \pm 1.51$ & 54.02 & $83.19 \pm 0.10$ & 87.50 \\
\bottomrule
\end{tabular}
\end{table}

A careful reading of the paper showed three places where my implementation did
not follow the text exactly: the paper picks the top-2 neighbours over the joint
fine-and-coarse graph (three prototypes instead of my six), its sample term has
no class-complement, and it says it uses the ConCutMix pipeline. I tested all
three. The first two each add about $0.6$ points, but this only brings SKCL to
the level of BCL ($50.92$ and $50.85$ against $50.85$). With the CutMix branch
added, SKCL reaches $50.66$ on CIFAR-100-LT, which is still $1.04$ below
ConCutMix. So the missing gain is not hidden in the details the paper leaves
open.

\subsection{SKCL on the authors' code reproduces the paper}

To rule out my own framework as the reason, I added SKCL to the official
ConCutMix code\footnote{\url{https://github.com/PanHaulin/ConCutMix}, commit
\texttt{f0e42f5}.} and kept its recipe unchanged. The published CIFAR-100
script does not run as it is: it does not set the number of classes, it needs
the cosine-classifier flag to give the prototype head the right shape, and one
indexing operation fails on recent PyTorch versions. After fixing these, the
official ConCutMix reaches $52.42$ at its best epoch after $200$ epochs, which
is $0.74$ below the published $53.16$.

\begin{figure}[t]
\centering
\includegraphics[width=0.78\textwidth]{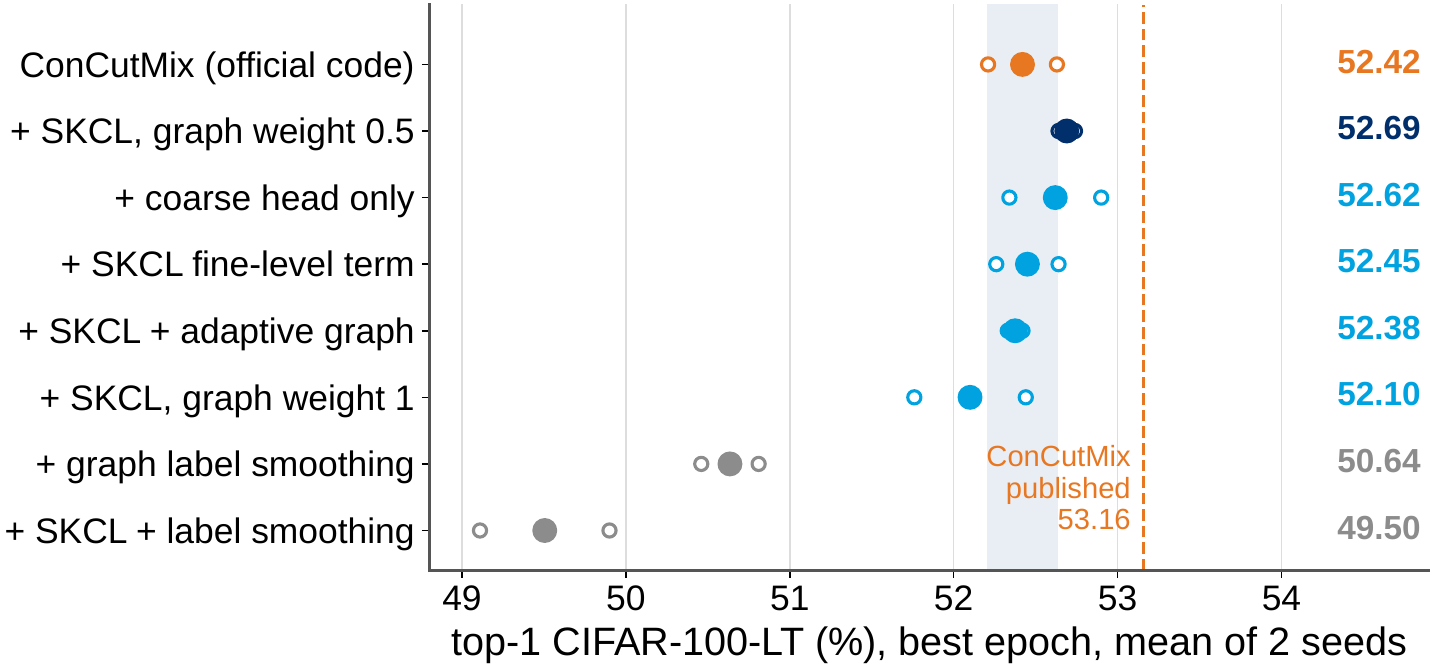}
\caption{SKCL variants added to the official ConCutMix code, CIFAR-100-LT,
$200$ epochs, best epoch. Filled markers: mean of two seeds; open markers:
single seeds; shaded band: the two ConCutMix seeds.}
\label{fig:official}
\end{figure}

On top of this code I added the parts of SKCL: a coarse classifier with its own
prototype head, the loss of Eq.~\ref{eq:skcl1} with the paper's member sets,
and, as options, graph label smoothing and the adaptive graph of
Section~\ref{sec:improve}. In a first stage I trained seven variants for $200$
epochs with two seeds (Figure~\ref{fig:official}). Five of them are within
$0.32$ points of ConCutMix. The two variants with graph label smoothing are
clearly worse overall ($-1.8$ and $-2.9$), because they move accuracy from the
frequent to the rare classes.

I then trained ConCutMix and the variant with half graph weight, the best
variant that uses the graph term, for $300$ epochs with three seeds
(Table~\ref{tab:official}). Here SKCL reaches $53.69 \pm 0.25$, only $0.33$
below the published $54.02$. It is $0.23$ points above ConCutMix, and the
difference is positive on all three seeds ($+0.05$, $+0.35$, $+0.28$). One
caveat is important. I chose this variant after looking at the first stage, so
the $+0.23$ is, if anything, optimistic. I had split the test set into two
fixed halves to keep selection and reporting apart, and on the half I did not
use for any decision the difference is $-0.31$, while it is $+0.73$ on the
other half.

\begin{table}[t]
\centering
\small
\caption{ConCutMix with and without SKCL on the official ConCutMix code,
CIFAR-100-LT, three seeds. Top-1 accuracy (\%).}
\label{tab:official}
\begin{tabular}{@{}lcc@{}}
\toprule
& Best epoch & Last-10 mean \\
\midrule
ConCutMix, 200 epochs (2 seeds) & $52.42 \pm 0.30$ & $52.20 \pm 0.27$ \\
ConCutMix, 300 epochs           & $53.46 \pm 0.38$ & $53.23 \pm 0.38$ \\
ConCutMix + SKCL, 300 epochs    & $53.69 \pm 0.25$ & $53.44 \pm 0.29$ \\
\midrule
SKCL, published (300 epochs)    & $54.02$ & \\
\bottomrule
\end{tabular}
\end{table}

\subsection{Where the claimed gain comes from}

\begin{figure}[t]
\centering
\includegraphics[width=0.58\textwidth]{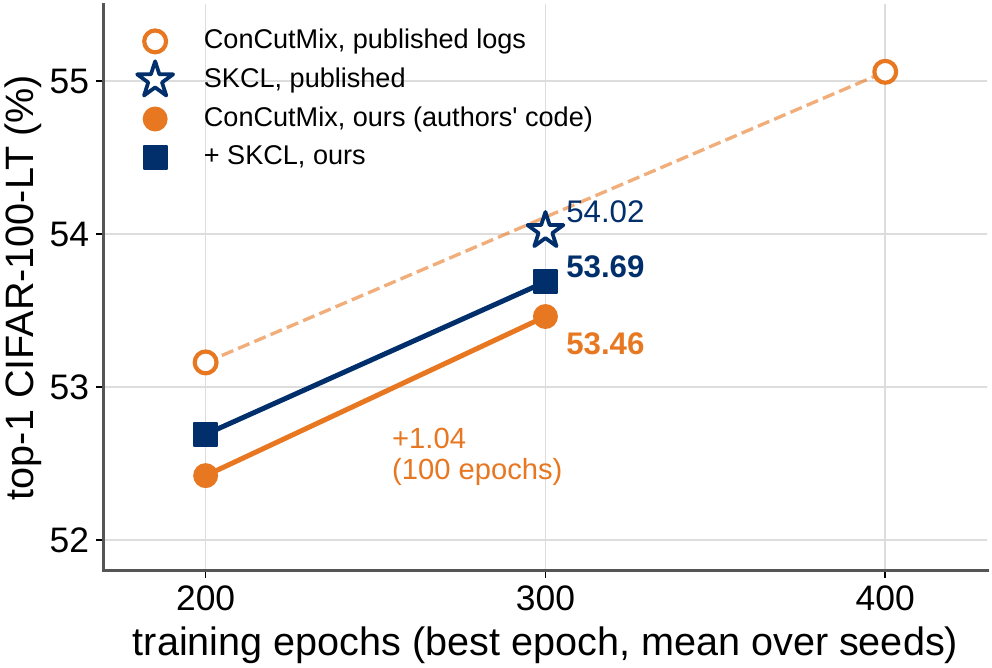}
\caption{Best-epoch accuracy against training epochs on CIFAR-100-LT. Open
markers: published values (official ConCutMix logs at $200$ and $400$ epochs,
SKCL paper at $300$). Filled markers: my runs of the official code.}
\label{fig:budget}
\end{figure}

The numbers above suggest a simple explanation. The SKCL paper trains for $300$
epochs and follows the ConCutMix pipeline, but its baseline rows are the
published $200$-epoch values. Table~\ref{tab:decomp} and
Figure~\ref{fig:budget} split the claimed $+2.01$ over BCL into three steps.
Going from BCL to ConCutMix gives $+1.15$ (published numbers). Going from $200$
to $300$ epochs gives $+1.04$ (measured on the official code). Adding the
semantic graph gives $+0.23$ (measured). The steps come from different runs, so
they are not exactly additive, but together they are enough to explain the
claimed gain, and the graph is clearly the smallest part.

\begin{table}[t]
\centering
\small
\caption{Breakdown of SKCL's claimed gain over BCL on CIFAR-100-LT (best
epoch, as in the original papers).}
\label{tab:decomp}
\begin{tabular}{@{}llr@{}}
\toprule
Step & Source & Gain \\
\midrule
BCL $\rightarrow$ ConCutMix & published ($52.01 \rightarrow 53.16$) & $+1.15$ \\
$200 \rightarrow 300$ epochs & measured, official code ($52.42 \rightarrow 53.46$) & $+1.04$ \\
+ semantic graph & measured, official code ($53.46 \rightarrow 53.69$) & $+0.23$ \\
\midrule
Sum & & $+2.42$ \\
Claimed by SKCL & published ($52.01 \rightarrow 54.02$) & $+2.01$ \\
\bottomrule
\end{tabular}
\end{table}

This also explains why my two settings first looked inconsistent. With the full
graph weight, the graph is slightly negative in both codebases ($-0.56$ against
BCL in mine, $-0.32$ against ConCutMix in theirs). Only the half-weight variant,
the best of seven, is slightly positive. All of these are close to zero.

\subsection{The graph is mostly redundant}

If the graph adds so little, is it a bad graph or a redundant one? These two
explanations predict different things. A bad graph should get better results
when it is improved. A redundant graph should not care which edges it contains.

The graph is not bad: its strongest links are correct and only $2\%$ of
classes have no neighbour with similarity above $0.4$. But it is redundant. For
a BCL model that never sees the graph, I compared each class's two most
confused classes with its two graph neighbours. They agree for $41.2\%$ of the
edges, against $2.0\%$ expected by chance, and the correlation between semantic
similarity and confusion over all class pairs is $r = 0.454$. Training with the
graph raises the overlap only to $44.3\%$. So the network already learns most of
these relations from the images.\footnote{These two numbers were measured on
checkpoints from before the bug fixes of Section~3 and not repeated.}

\begin{figure}[t]
\centering
\includegraphics[width=0.85\textwidth]{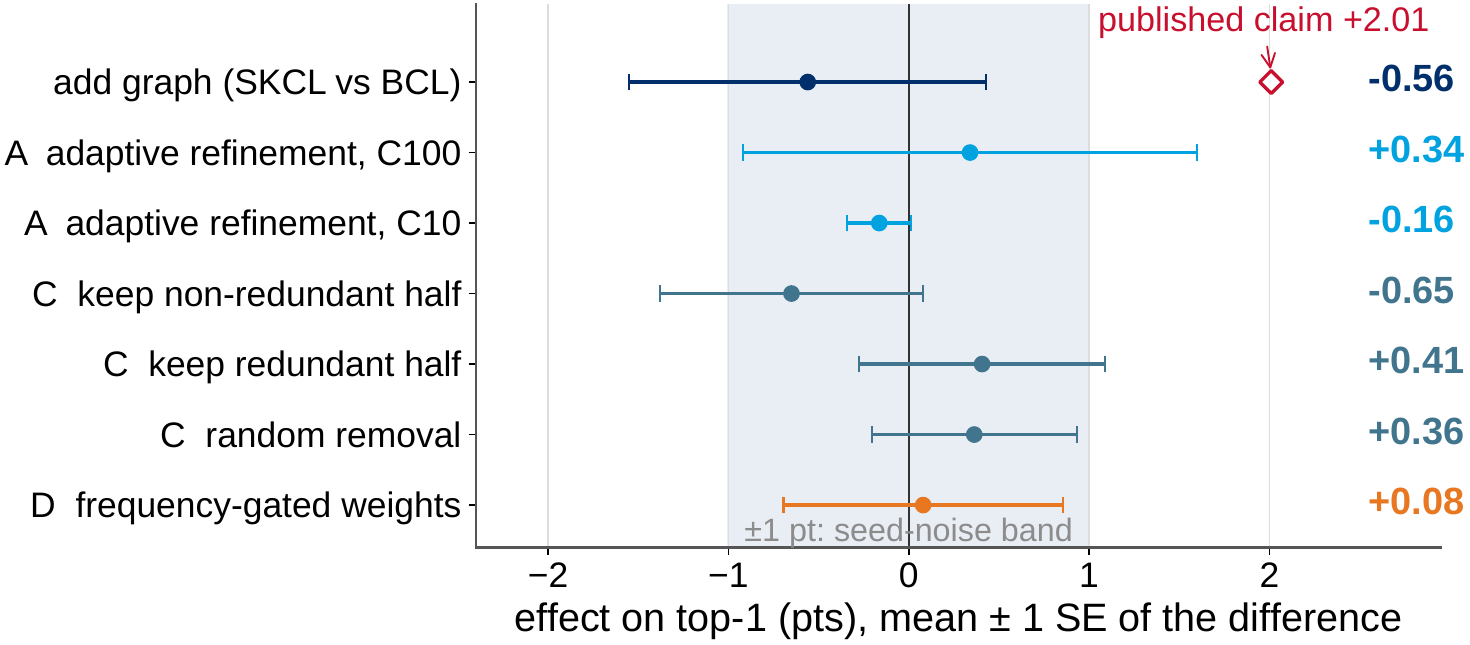}
\caption{Effect of the graph and of interventions on it, CIFAR-100-LT (and
CIFAR-10-LT where marked), mean $\pm$ 1 SE of the difference over three seeds.
The shaded band marks $\pm 1$ point of seed noise; the red marker is the
published claim.}
\label{fig:null}
\end{figure}

I then changed the graph in several ways (Figure~\ref{fig:null}). Keeping only
the edges the model does not already know, only the edges it does know, or a
random subset of the same size gives results between $50.38$ and $51.44$. Two
retrainings on the \emph{same} graph already differ by $0.74$, so all of these
are within noise. Weighting neighbours by class frequency also changes nothing
measurable. If the loss used the identity of the neighbours, these changes
should matter. That they do not suggests that the graph works more like a
general regulariser on the prototypes than as a source of class-specific
knowledge.

\subsection{Improving SKCL}
\label{sec:improve}

I also tried to make SKCL better. Table~\ref{tab:improve} lists every change
that improved it in at least one setting.

\begin{table}[t]
\centering
\small
\caption{Changes that improve SKCL. Own: my framework, last-10 mean, three
seeds. Official: official ConCutMix code, best epoch, $200$ epochs, two seeds.}
\label{tab:improve}
\begin{tabular}{@{}lllr@{}}
\toprule
Change & Setting & Reference $\rightarrow$ changed & Gain \\
\midrule
+ CutMix branch            & own, C100      & $50.29 \rightarrow 51.35$ & $+1.06$ \\
+ CutMix branch            & own, C10       & $83.19 \rightarrow 84.13$ & $+0.94$ \\
paper's member sets        & own, C100      & $50.29 \rightarrow 50.92$ & $+0.63$ \\
adaptive graph             & own, C100      & $50.29 \rightarrow 50.63$ & $+0.34$ \\
adaptive graph             & official, C100 & $52.10 \rightarrow 52.38$ & $+0.28$ \\
graph weight $0.5$ instead of $1$ & official, C100 & $52.10 \rightarrow 52.69$ & $+0.59$ \\
graph label smoothing, Few classes & official, C100 & $40.1 \rightarrow 44.5$ & $+4.4$ \\
\bottomrule
\end{tabular}
\end{table}

The largest gain comes from combining SKCL with the CutMix branch of ConCutMix.
On CIFAR-10-LT this combination ($84.13$) is better than both BCL ($83.53$) and
ConCutMix ($83.84$). This fits the breakdown above, because again the gain comes
from the classification pipeline and not from the graph. Halving the graph
weight also helps, which suggests that the graph term is too strong at weight
one.

The \textbf{adaptive graph} is my own extension. The language graph is built
before the network sees any image, but what makes classes related for a
classifier is whether they look similar. So I keep a running average of each
class's embedding, and every ten epochs after a warm-up of twenty epochs I mix
the language graph $50/50$ with the similarities of these class means and pick
the neighbours again. I always mix with the original language graph, because a
graph built only from the network's own embeddings would just confirm what the
network already believes. About $40\%$ of the top-2 edges change at the first
refresh, so the network's idea of ``related'' is clearly different from the
language model's. Still, accuracy changes only a little: $+0.34$ in my framework
and $+0.28$ on the official code, but $-0.16$ on CIFAR-10-LT. With seed SDs of
up to $1.5$ points, I would call this a small and consistent trend, not an
established improvement. It also fits the redundancy finding: if rewiring
$40\%$ of the edges hardly matters, the specific edges are not where the effect
comes from.

Graph label smoothing, which moves $10\%$ of each training label to the two
graph neighbours, is a trade-off rather than an improvement. It raises the Few
group by $4.4$ points, but costs around ten points on the Many group.

\section{Discussion and Limitations}

The main message of this study is that SKCL's number can be reproduced, but its
explanation does not hold up well. On the ConCutMix pipeline with $300$ epochs I
get within $0.33$ points of the published result. But the semantic graph itself
contributes only about $0.2$ points, and most of the reported gain over BCL comes
from the CutMix branch and the longer training. On CIFAR, the reason is that the
graph mostly tells the network things it learns anyway.

This has two practical consequences. First, for methods whose input is the
output of a language model, that output has to be published. The graph depends
strongly on how it is built: during development, a simple TF-IDF embedding of
the same descriptions linked \textsf{otter} to \textsf{flowers}, while MiniLM
links it to \textsf{seal} and \textsf{lobster}. Second, methods should be
compared at the same training budget. A $100$-epoch difference is worth about
one point here, which is more than most of the differences between methods.

The study has clear limitations. The class descriptions are mine, and some
hyperparameters of SKCL ($k$, the coarse weight, $\tau'$) are not given in the
paper, so I chose them. Three seeds can only resolve differences of about one
point, and less for the Few group. On the official code, the first stage uses
two seeds, and the second-stage variant was chosen after looking at the
results. The redundancy numbers come from checkpoints before the bug fixes.
Finally, everything here is on CIFAR at $32 \times 32$ pixels with $100$ very
different classes. On fine-grained datasets such as CUB-200~\cite{wah2011cub},
iNaturalist~\cite{vanhorn2018inat} or ImageNet-LT~\cite{liu2019oltr}, language
may know things that are hard to see, and the graph could be much less
redundant. Measuring the redundancy there first would show whether SKCL can
help in that setting.

\section{Conclusion}

I reproduced SKCL, BCL and ConCutMix in one framework and on the authors'
ConCutMix code. Built on BCL, SKCL does not improve over BCL. Built on ConCutMix
with $300$ epochs, it comes within $0.33$ points of the published result, but
the semantic graph adds only about $0.2$ points at equal budget, and most of the
claimed gain is explained by the CutMix branch and the longer schedule. A model
that never sees the graph already knows most of it. Some changes improve SKCL,
mainly the CutMix branch, while the adaptive graph gives small gains within
noise. I hope the detailed record of differences, bugs and fixes is also useful
for others who try to reproduce methods without released code.

\paragraph{Code and data.} Code, class descriptions, graphs and run scripts:
\url{https://github.com/sushrutghimire1/Reproduction-and-extension-of-SKCL}.

\paragraph{Acknowledgements.} This work started as a seminar project in the
Representation Learning seminar of the IDEA Lab at FAU. I thank Prof.\ Dr.\
Bernhard Kainz and Mischa Dombrowski for supervising it. Experiments were run on
the TinyGPU cluster of NHR@FAU.

\paragraph{Use of AI tools.} The class descriptions were generated with an LLM
(Claude, Anthropic), as described in Section~3. Generative AI tools were also
used to assist with code and writing. The author checked all content and takes
full responsibility for it.


\appendix
\FloatBarrier
\section{Per-Seed Results}

Table~\ref{tab:perseed} lists every run of my framework after the fixes (\texttt{FIX\_bcl\_c100\_s0\_dup} is the duplicate-run check and is not used in any mean), and Table~\ref{tab:perseed-f} lists the runs on the official ConCutMix code.

\begin{table}[p]
\centering
\scriptsize
\caption[Per-seed results of all corrected-pipeline runs]{Per-seed results of
all corrected-pipeline runs, $\beta = 100$. Last-10: mean top-1 over the final
ten epochs (the reported metric); Best: maximum top-1 over all epochs (shown for
comparison only); Many, Med.\ (Medium) and Few are last-10 group accuracies.
C100/C10: CIFAR-100-LT/CIFAR-10-LT, which has no Few classes at $\beta = 100$.
Run-name prefixes: bcl = BCL, ccm = ConCutMix, ccm\_g0 = ConCutMix
with $\gamma = 0$, skcl = SKCL, adapt = SKCL + adaptive graph,
cutmixcombo(\_nowt) = frequency-gated (binary) SKCL graph + CutMix,
full/surprising/redundant/random = graph-filtering variants; PF\_joint, PF\_nocc, PF\_graph, PF\_full = paper-faithful paper-text variants: joint member sets, no class-complement, both, both + CutMix branch; s$n$ = seed $n$.}
\label{tab:perseed}
\begin{tabular}{@{}llccccc@{}}
\toprule
Run & Data & Last-10 & Best & Many & Med. & Few \\
\midrule
\texttt{FIX\_adapt\_c100\_s0} & C100 & 51.34 & 51.60 & 68.77 & 51.54 & 30.78 \\
\texttt{FIX\_adapt\_c100\_s1} & C100 & 48.83 & 49.09 & 65.95 & 49.25 & 28.37 \\
\texttt{FIX\_adapt\_c100\_s2} & C100 & 51.72 & 51.84 & 66.65 & 52.05 & 33.92 \\
\texttt{FIX\_adapt\_c10\_s0} & C10 & 83.14 & 83.72 & 85.03 & 75.56 & --- \\
\texttt{FIX\_adapt\_c10\_s1} & C10 & 82.70 & 83.14 & 84.64 & 74.95 & --- \\
\texttt{FIX\_adapt\_c10\_s2} & C10 & 83.25 & 83.98 & 84.87 & 76.79 & --- \\
\texttt{FIX\_bcl\_c100\_s0} & C100 & 50.26 & 50.59 & 67.24 & 51.89 & 28.56 \\
\texttt{FIX\_bcl\_c100\_s0\_dup} & C100 & 50.26 & 50.59 & 67.24 & 51.89 & 28.56 \\
\texttt{FIX\_bcl\_c100\_s1} & C100 & 50.52 & 50.66 & 67.59 & 50.71 & 30.36 \\
\texttt{FIX\_bcl\_c100\_s2} & C100 & 51.77 & 52.08 & 67.48 & 51.91 & 33.27 \\
\texttt{FIX\_bcl\_c10\_s0} & C10 & 83.27 & 83.70 & 84.98 & 76.43 & --- \\
\texttt{FIX\_bcl\_c10\_s1} & C10 & 83.65 & 83.86 & 85.64 & 75.72 & --- \\
\texttt{FIX\_bcl\_c10\_s2} & C10 & 83.68 & 84.20 & 85.34 & 77.04 & --- \\
\texttt{FIX\_ccm\_c100\_s0} & C100 & 51.61 & 51.92 & 66.91 & 52.59 & 32.62 \\
\texttt{FIX\_ccm\_c100\_s1} & C100 & 50.96 & 51.14 & 66.39 & 51.83 & 31.95 \\
\texttt{FIX\_ccm\_c100\_s2} & C100 & 52.55 & 52.88 & 66.80 & 52.99 & 35.42 \\
\texttt{FIX\_ccm\_c10\_s0} & C10 & 84.29 & 84.39 & 85.32 & 80.15 & --- \\
\texttt{FIX\_ccm\_c10\_s1} & C10 & 83.48 & 84.03 & 84.66 & 78.73 & --- \\
\texttt{FIX\_ccm\_c10\_s2} & C10 & 83.74 & 83.93 & 85.01 & 78.64 & --- \\
\texttt{FIX\_ccm\_g0\_c100\_s0} & C100 & 51.40 & 51.52 & 66.85 & 52.18 & 32.45 \\
\texttt{FIX\_ccm\_g0\_c100\_s1} & C100 & 50.92 & 50.99 & 66.43 & 50.72 & 33.05 \\
\texttt{FIX\_ccm\_g0\_c100\_s2} & C100 & 52.90 & 52.98 & 65.29 & 53.78 & 37.41 \\
\texttt{FIX\_cutmixcombo\_c100\_s0} & C100 & 50.96 & 51.24 & 66.35 & 51.90 & 31.90 \\
\texttt{FIX\_cutmixcombo\_c100\_s1} & C100 & 50.58 & 50.84 & 67.01 & 49.52 & 32.66 \\
\texttt{FIX\_cutmixcombo\_c100\_s2} & C100 & 52.51 & 52.78 & 67.17 & 52.68 & 35.22 \\
\texttt{FIX\_cutmixcombo\_nowt\_c100\_s0} & C100 & 51.90 & 52.00 & 66.03 & 53.53 & 33.50 \\
\texttt{FIX\_cutmixcombo\_nowt\_c100\_s1} & C100 & 50.28 & 50.52 & 66.85 & 50.25 & 30.97 \\
\texttt{FIX\_cutmixcombo\_nowt\_c100\_s2} & C100 & 51.63 & 51.98 & 66.26 & 51.08 & 35.20 \\
\texttt{FIX\_full\_s0} & C100 & 50.84 & 51.00 & 67.39 & 51.18 & 31.12 \\
\texttt{FIX\_full\_s1} & C100 & 50.52 & 50.76 & 67.19 & 49.79 & 31.93 \\
\texttt{FIX\_full\_s2} & C100 & 51.74 & 52.01 & 67.15 & 51.22 & 34.38 \\
\texttt{FIX\_random\_s0} & C100 & 51.42 & 51.54 & 68.13 & 50.64 & 32.82 \\
\texttt{FIX\_random\_s1} & C100 & 50.63 & 50.78 & 68.07 & 50.18 & 30.82 \\
\texttt{FIX\_random\_s2} & C100 & 52.14 & 52.27 & 67.05 & 51.09 & 35.98 \\
\texttt{FIX\_redundant\_s0} & C100 & 51.79 & 51.88 & 67.91 & 51.74 & 33.05 \\
\texttt{FIX\_redundant\_s1} & C100 & 50.31 & 50.52 & 66.34 & 49.54 & 32.52 \\
\texttt{FIX\_redundant\_s2} & C100 & 52.22 & 52.32 & 67.57 & 51.14 & 35.56 \\
\texttt{FIX\_skcl\_c100\_s0} & C100 & 50.02 & 50.11 & 66.69 & 50.54 & 29.96 \\
\texttt{FIX\_skcl\_c100\_s1} & C100 & 48.93 & 49.01 & 66.31 & 47.75 & 30.04 \\
\texttt{FIX\_skcl\_c100\_s2} & C100 & 51.92 & 52.22 & 66.82 & 51.27 & 35.32 \\
\texttt{FIX\_skcl\_c10\_s0} & C10 & 83.15 & 83.44 & 85.30 & 74.59 & --- \\
\texttt{FIX\_skcl\_c10\_s1} & C10 & 83.31 & 83.49 & 85.09 & 76.23 & --- \\
\texttt{FIX\_skcl\_c10\_s2} & C10 & 83.12 & 83.63 & 84.96 & 75.77 & --- \\
\texttt{FIX\_surprising\_s0} & C100 & 50.16 & 50.32 & 66.47 & 50.31 & 30.95 \\
\texttt{FIX\_surprising\_s1} & C100 & 49.42 & 49.57 & 67.35 & 49.14 & 28.83 \\
\texttt{FIX\_surprising\_s2} & C100 & 51.57 & 51.78 & 66.57 & 50.41 & 35.41 \\
\midrule
\texttt{PF\_full\_c100\_s0} & C100 & 50.24 & 50.41 & 65.85 & 50.28 & 31.97 \\
\texttt{PF\_full\_c100\_s1} & C100 & 49.77 & 49.91 & 66.74 & 49.12 & 30.74 \\
\texttt{PF\_full\_c100\_s2} & C100 & 51.98 & 52.12 & 65.77 & 51.63 & 36.30 \\
\texttt{PF\_full\_c10\_s0} & C10 & 84.61 & 84.77 & 85.49 & 81.10 & --- \\
\texttt{PF\_full\_c10\_s1} & C10 & 83.89 & 84.67 & 84.59 & 81.11 & --- \\
\texttt{PF\_full\_c10\_s2} & C10 & 83.89 & 84.30 & 84.23 & 82.53 & --- \\
\texttt{PF\_graph\_c100\_s0} & C100 & 50.04 & 50.39 & 67.31 & 50.14 & 29.79 \\
\texttt{PF\_graph\_c100\_s1} & C100 & 49.61 & 49.72 & 67.65 & 48.19 & 30.23 \\
\texttt{PF\_graph\_c100\_s2} & C100 & 51.64 & 51.71 & 66.40 & 51.54 & 34.53 \\
\texttt{PF\_joint\_c100\_s0} & C100 & 50.83 & 50.94 & 66.52 & 51.89 & 31.29 \\
\texttt{PF\_joint\_c100\_s1} & C100 & 49.83 & 50.04 & 66.75 & 49.22 & 30.79 \\
\texttt{PF\_joint\_c100\_s2} & C100 & 52.11 & 52.29 & 67.37 & 52.47 & 33.89 \\
\texttt{PF\_nocc\_c100\_s0} & C100 & 50.43 & 50.59 & 67.45 & 50.94 & 30.00 \\
\texttt{PF\_nocc\_c100\_s1} & C100 & 50.07 & 50.32 & 67.45 & 49.33 & 30.66 \\
\texttt{PF\_nocc\_c100\_s2} & C100 & 52.05 & 52.28 & 67.99 & 52.11 & 33.39 \\
\bottomrule
\end{tabular}
\end{table}

\begin{table}[h]
\centering
\scriptsize
\caption[Per-seed results on the official ConCutMix code]{Per-seed results on the official ConCutMix code
(official ConCutMix code, CIFAR-100-LT, $\beta = 100$). Best: best-epoch
top-1 on the full test set; Last-10: last-10 mean; A/B: last-10 mean on the two
test halves; Many, Med., Few: last-10 group accuracies. OCCM = ConCutMix;
SK\_ = ConCutMix + SKCL variant (full: coarse head, joint member
sets, $w = 1$; full05: $w = 0.5$; coarse: coarse head only; fine: fine-level
Term~1; adapt: full + adaptive graph; ls: graph label smoothing; all: full +
ls); 200/300: epochs; s$n$: seed.}
\label{tab:perseed-f}
\begin{tabular}{@{}lccccccc@{}}
\toprule
Run & Best & Last-10 & A & B & Many & Med. & Few \\
\midrule
\texttt{OCCM200\_s0} & 52.63 & 52.39 & 52.69 & 52.08 & 64.91 & 49.91 & 40.68 \\
\texttt{OCCM200\_s1} & 52.21 & 52.01 & 52.19 & 51.84 & 64.45 & 50.22 & 39.60 \\
\texttt{SK\_full05\_200\_s0} & 52.74 & 52.61 & 52.32 & 52.90 & 64.68 & 50.29 & 41.25 \\
\texttt{SK\_full05\_200\_s1} & 52.64 & 52.47 & 52.22 & 52.71 & 65.60 & 49.27 & 40.87 \\
\texttt{SK\_coarse\_200\_s0} & 52.90 & 52.54 & 52.84 & 52.25 & 65.14 & 49.51 & 41.39 \\
\texttt{SK\_coarse\_200\_s1} & 52.34 & 52.17 & 52.90 & 51.44 & 64.53 & 50.12 & 40.13 \\
\texttt{SK\_fine\_200\_s0} & 52.26 & 51.95 & 52.40 & 51.50 & 63.60 & 49.22 & 41.54 \\
\texttt{SK\_fine\_200\_s1} & 52.64 & 52.38 & 52.96 & 51.80 & 64.29 & 50.84 & 40.29 \\
\texttt{SK\_adapt\_200\_s0} & 52.42 & 52.08 & 51.99 & 52.18 & 64.46 & 49.64 & 40.50 \\
\texttt{SK\_adapt\_200\_s1} & 52.33 & 52.10 & 52.72 & 51.48 & 64.57 & 49.58 & 40.49 \\
\texttt{SK\_full\_200\_s0} & 51.76 & 51.50 & 51.56 & 51.45 & 64.31 & 48.41 & 40.18 \\
\texttt{SK\_full\_200\_s1} & 52.44 & 52.33 & 52.73 & 51.93 & 65.29 & 49.70 & 40.26 \\
\texttt{SK\_ls\_200\_s0} & 50.81 & 50.20 & 50.83 & 49.58 & 54.94 & 49.94 & 44.99 \\
\texttt{SK\_ls\_200\_s1} & 50.46 & 49.97 & 49.92 & 50.01 & 55.27 & 49.76 & 44.02 \\
\texttt{SK\_all\_200\_s0} & 49.11 & 48.79 & 48.49 & 49.09 & 54.48 & 47.44 & 43.72 \\
\texttt{SK\_all\_200\_s1} & 49.90 & 49.13 & 49.11 & 49.15 & 54.69 & 48.07 & 43.87 \\
\midrule
\texttt{OCCM300\_s0} & 53.79 & 53.54 & 53.97 & 53.11 & 66.98 & 52.01 & 39.64 \\
\texttt{OCCM300\_s1} & 53.05 & 52.81 & 53.29 & 52.34 & 65.89 & 51.35 & 39.27 \\
\texttt{OCCM300\_s2} & 53.54 & 53.35 & 53.68 & 53.02 & 66.90 & 51.55 & 39.64 \\
\texttt{SK\_full05\_300\_s0} & 53.84 & 53.52 & 52.97 & 54.07 & 66.25 & 51.18 & 41.39 \\
\texttt{SK\_full05\_300\_s1} & 53.40 & 53.12 & 53.38 & 52.87 & 66.49 & 50.79 & 40.25 \\
\texttt{SK\_full05\_300\_s2} & 53.82 & 53.69 & 53.66 & 53.73 & 67.07 & 52.12 & 39.92 \\
\bottomrule
\end{tabular}
\end{table}


\begin{thebibliography}{99}
\small

\bibitem{liu2025skcl}
Y.~Liu, L.~Yang, and Y.~Wang.
\newblock Long-Tailed Classification with Multi-Granularity Semantics.
\newblock In \emph{ICCV}, 2025.

\bibitem{zhu2022bcl}
J.~Zhu, Z.~Wang, J.~Chen, Y.-P.~P. Chen, and Y.-G. Jiang.
\newblock Balanced Contrastive Learning for Long-Tailed Visual Recognition.
\newblock In \emph{CVPR}, 2022.

\bibitem{pan2024concutmix}
H.~Pan, Y.~Guo, Q.~Deng, H.~Yang, J.~Chen, and Y.~Chen.
\newblock Enhanced Long-Tailed Recognition with Contrastive CutMix
  Augmentation.
\newblock \emph{IEEE Transactions on Image Processing}, 2024.

\bibitem{cao2019ldam}
K.~Cao, C.~Wei, A.~Gaidon, N.~Arechiga, and T.~Ma.
\newblock Learning Imbalanced Datasets with Label-Distribution-Aware Margin
  Loss.
\newblock In \emph{NeurIPS}, 2019.

\bibitem{khosla2020supcon}
P.~Khosla, P.~Teterwak, C.~Wang, A.~Sarna, Y.~Tian, P.~Isola, A.~Maschinot,
  C.~Liu, and D.~Krishnan.
\newblock Supervised Contrastive Learning.
\newblock In \emph{NeurIPS}, 2020.

\bibitem{chen2020simclr}
T.~Chen, S.~Kornblith, M.~Norouzi, and G.~Hinton.
\newblock A Simple Framework for Contrastive Learning of Visual
  Representations.
\newblock In \emph{ICML}, 2020.

\bibitem{menon2021logit}
A.~K. Menon, S.~Jayasumana, A.~S. Rawat, H.~Jain, A.~Veit, and S.~Kumar.
\newblock Long-Tail Learning via Logit Adjustment.
\newblock In \emph{ICLR}, 2021.

\bibitem{yun2019cutmix}
S.~Yun, D.~Han, S.~J. Oh, S.~Chun, J.~Choe, and Y.~Yoo.
\newblock CutMix: Regularization Strategy to Train Strong Classifiers with
  Localizable Features.
\newblock In \emph{ICCV}, 2019.

\bibitem{devries2017cutout}
T.~DeVries and G.~W. Taylor.
\newblock Improved Regularization of Convolutional Neural Networks with Cutout.
\newblock \emph{arXiv:1708.04552}, 2017.

\bibitem{cubuk2019autoaugment}
E.~D. Cubuk, B.~Zoph, D.~Man\'e, V.~Vasudevan, and Q.~V. Le.
\newblock AutoAugment: Learning Augmentation Strategies from Data.
\newblock In \emph{CVPR}, 2019.

\bibitem{reimers2019sbert}
N.~Reimers and I.~Gurevych.
\newblock Sentence-BERT: Sentence Embeddings using Siamese BERT-Networks.
\newblock In \emph{EMNLP}, 2019.

\bibitem{he2016resnet}
K.~He, X.~Zhang, S.~Ren, and J.~Sun.
\newblock Deep Residual Learning for Image Recognition.
\newblock In \emph{CVPR}, 2016.

\bibitem{krizhevsky2009cifar}
A.~Krizhevsky.
\newblock Learning Multiple Layers of Features from Tiny Images.
\newblock Technical report, University of Toronto, 2009.

\bibitem{wah2011cub}
C.~Wah, S.~Branson, P.~Welinder, P.~Perona, and S.~Belongie.
\newblock The Caltech-UCSD Birds-200-2011 Dataset.
\newblock Technical Report CNS-TR-2011-001, Caltech, 2011.

\bibitem{vanhorn2018inat}
G.~Van~Horn, O.~Mac~Aodha, Y.~Song, Y.~Cui, C.~Sun, A.~Shepard, H.~Adam,
  P.~Perona, and S.~Belongie.
\newblock The iNaturalist Species Classification and Detection Dataset.
\newblock In \emph{CVPR}, 2018.

\bibitem{liu2019oltr}
Z.~Liu, Z.~Miao, X.~Zhan, J.~Wang, B.~Gong, and S.~X. Yu.
\newblock Large-Scale Long-Tailed Recognition in an Open World.
\newblock In \emph{CVPR}, 2019.

\end{thebibliography}
\end{document}